\documentclass[10pt,twocolumn]{article} 
\usepackage{simpleConference}
\usepackage{times}
\usepackage{graphicx}
\usepackage{amssymb}
\usepackage{url}
\usepackage{latexsym}
\usepackage{caption}
\usepackage{subcaption}

\newcommand\blfootnote[1]{%
  \begingroup
  \renewcommand\thefootnote{}\footnote{#1}%
  \addtocounter{footnote}{-1}%
  \endgroup
}

\begin{document}

\title{Iris Presentation Attack Detection: Where Are We Now?}

\author{Aidan Boyd*, Zhaoyuan Fang*, Adam Czajka, and Kevin W. Bowyer\\
University of Notre Dame\\
{\tt\small \{aboyd3, zfang, aczajka, kwb\}@nd.edu}
}

\maketitle
\thispagestyle{empty}
\blfootnote{* indicates equal contribution}

\begin{abstract}
As the popularity of iris recognition systems increases, the importance of effective security measures against presentation attacks becomes paramount. 
This work presents an overview of the most important advances in the area of iris presentation attack detection published in the recent two years.
Newly-released, publicly-available datasets for development and evaluation of iris presentation attack detection are discussed. Recent literature can be seen to be broken into three categories: traditional ``hand-crafted'' feature extraction and classification, deep learning-based solutions, and hybrid approaches fusing both methodologies. Conclusions of modern approaches underscore the difficulty of this task. Finally, commentary on possible directions for future research is provided.
\end{abstract}




\section{Introduction}

Iris recognition has gained a place as one of the fastest and most secure biometric authentication methods. It has proven effective in many large-scale applications such as national identification (\cite{AADHAAR}) and border control (\cite{NEXUS}). With the increased deployment, the security of these systems against attacks becomes critical. The most common form of security breach
is {\it presentation attacks}. This term refers to a sample being presented to an iris sensor with the goal of manipulating the biometric system into an incorrect decision. 

Presentation attack samples can be used to either {\it impersonate} an identity or to {\it conceal} an identity. Impostor Attack Presentation is the term used for impersonation attacks, while Concealer Attack Presentation describes an attack meant to hide the user's identity. Users can also attempt to enroll with a presentation attack sample to continually manipulate the system.

Researchers must develop systems that are robust to some or all of the aforementioned attacks, and {\it Presentation Attack Detection} (PAD) -- the term coined during one of the ISO/IEC SC37 meetings (\cite{ISO_30107_1_2016}) -- is the area of research aiming at creating biometric systems that can determine whether a sample presented to a sensor is from a {\it bona fide} iris or is a presentation attack. 
This goal is difficult to achieve due to the ever-changing attack landscape. As systems become more resilient to known attack types, new attacks are being formulated and deployed. This survey reviews studies relating to both closed-set PAD, where the testing attack types are known during training, and open-set PAD, where the testing attack types are unknown during training.

Solutions to iris PAD can be either software-based and hardware-based. Software-based solutions use only the information present in the image to make the classification, whereas hardware solutions employ additional illumination or sensors to aid the classification. One can also see effective combinations of those two approached to strengthen the PAD capabilities. This survey discusses mainly software-based solutions, however, some recent advances in hardware solutions are also discussed. 

This work builds upon a comprehensive iris PAD survey by \cite{Czajka2018_ACM_Survey} and summarizes the most important developments in the field since June 2018. 
In Section 2 the common terminology and types of attack instrumentation are explained. Section 3 outlines the current publicly available PAD datasets. Section 4 presents the most recent works in iris PAD and in Section 5 the performance of these methods are discussed. Future research directions are given in section 6 and the work is summarized and concluded in Section 7.

\begin{figure*}
     \centering
     \begin{subfigure}[b]{0.19\textwidth}
         \centering
         \includegraphics[width=1\textwidth]{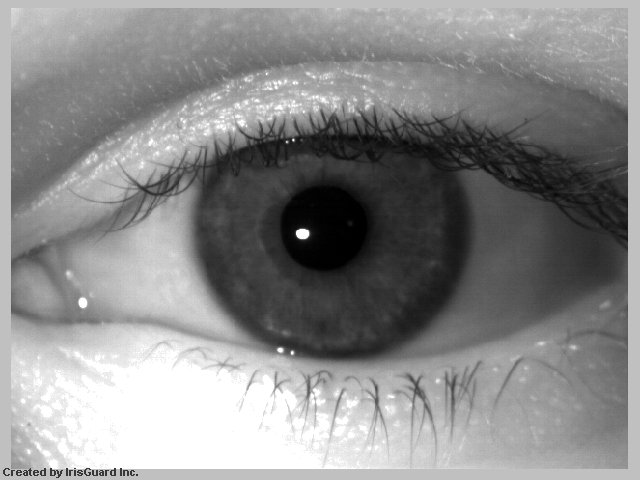}
     \end{subfigure}
     \begin{subfigure}[b]{0.19\textwidth}
         \centering
         \includegraphics[width=1\textwidth]{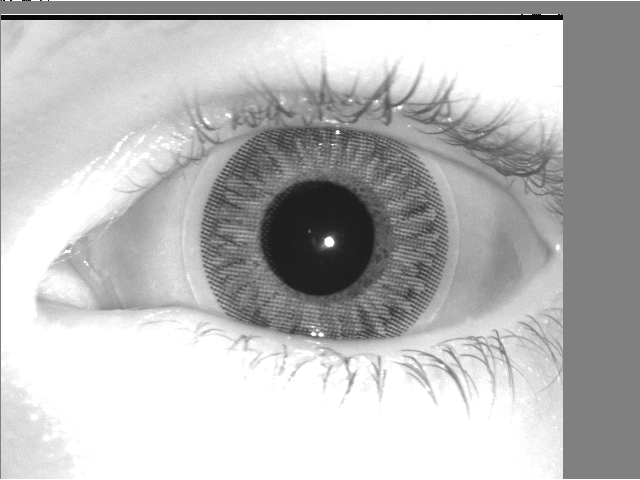}
     \end{subfigure}
     \begin{subfigure}[b]{0.19\textwidth}
         \centering
         \includegraphics[width=1\textwidth]{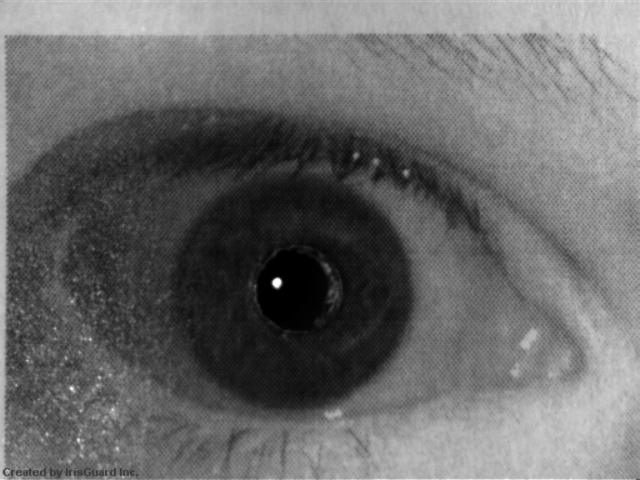}
     \end{subfigure}
     \begin{subfigure}[b]{0.19\textwidth}
         \centering
         \includegraphics[width=1\textwidth]{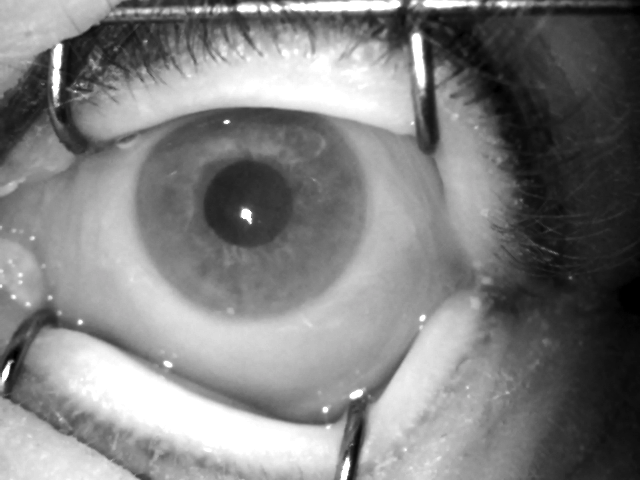}
     \end{subfigure}
     \begin{subfigure}[b]{0.19\textwidth}
         \centering
         \includegraphics[width=1\textwidth]{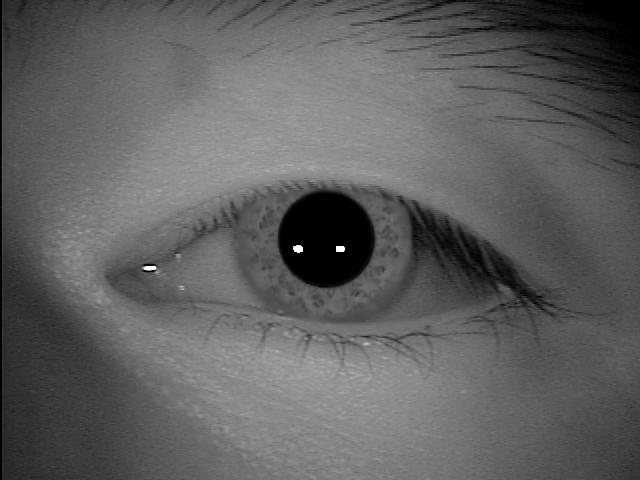}
     \end{subfigure}
     \bigskip
          \centering
     \begin{subfigure}[b]{0.19\textwidth}
         \centering
         \includegraphics[width=1\textwidth]{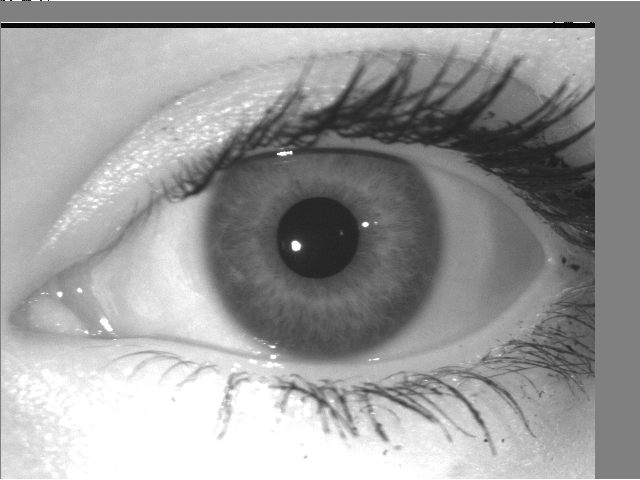}
         \caption{Bona fide Irises}
         \label{subfig:auth}
     \end{subfigure}
     \begin{subfigure}[b]{0.19\textwidth}
         \centering
         \includegraphics[width=1\textwidth]{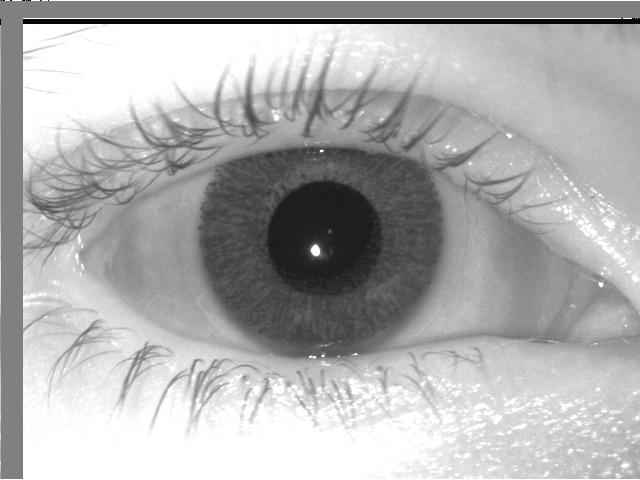}
         \caption{Textured Contact Lenses}
         \label{subfig:text}
     \end{subfigure}
     \begin{subfigure}[b]{0.19\textwidth}
         \centering
         \includegraphics[width=1\textwidth]{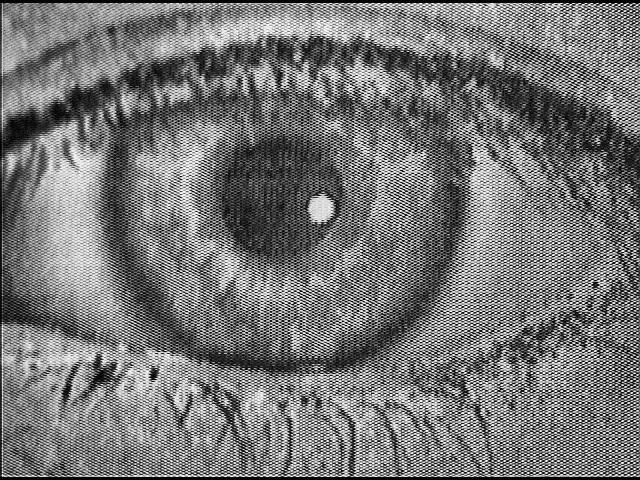}
         \caption{Paper Printouts}
         \label{subfig:print}
     \end{subfigure}
     \begin{subfigure}[b]{0.19\textwidth}
         \centering
         \includegraphics[width=1\textwidth]{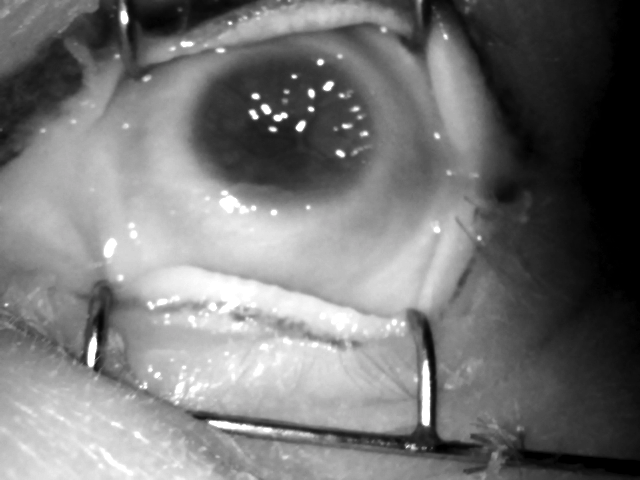}
         \caption{Post-mortem Samples}
         \label{subfig:post}
     \end{subfigure}
     \begin{subfigure}[b]{0.19\textwidth}
         \centering
         \includegraphics[width=1\textwidth]{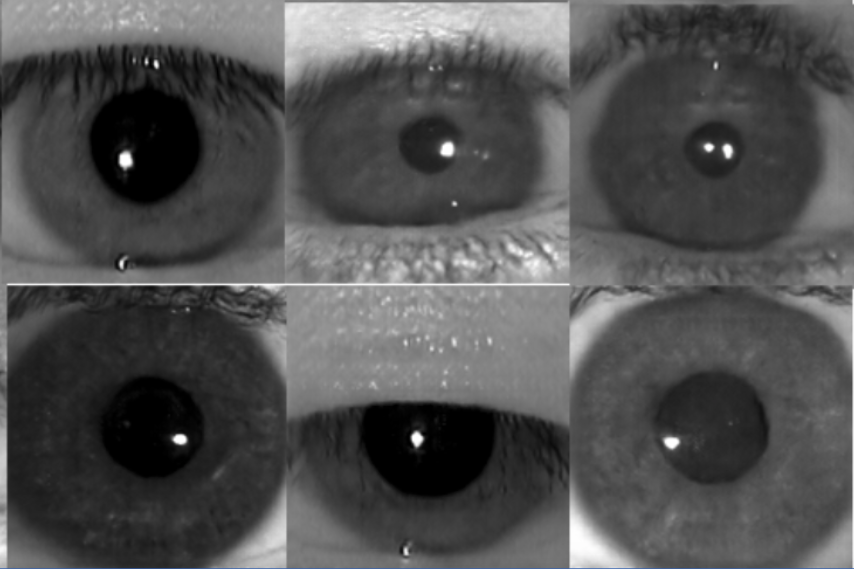}
         \caption{Synthetic Irises}
         \label{subfig:synth}
     \end{subfigure}
    \label{fig:attack_types}
    \vskip-5mm
    \caption{Illustration of live iris images, compliant with ISO/IEC 19794-6 (a), and presentation attack instruments popular in research databases (b-e). The top (d) image is an early-stage post-mortem sample whereas the bottom sample shown in (d) represents a later-stage capture. The example shown in top (e) picture illustrates the generation of synthetic iris texture by combining bona fide iris texture patches to form a new iris texture. The examples shown in bottom (e) section illustrate synthetic irises generated by a Generative Adversarial Network (GAN) developed by \cite{Yadava}. Note that some departures from a live iris are easy to observe, such as an extra pattern overlaid on the actual iris tissue (printouts and textured contact lenses) or metal retractors (post-mortem cases). It may be, however, harder to judge on the authenticity of an iris image in case of good-quality synthetic samples and newly deceased post-mortem samples. }
\end{figure*}

\section{Terminology and Attack Instrumentation}

\cite{Czajka2018_ACM_Survey} followed the vocabulary recommended in ISO/IEC 30107-3:2017. Here, we follow the same practice and provide a review of the PAD terminology.

\subsection{Presentation Attack Instruments}

A presentation to a biometric sensor is either a \textit{bona fide} presentation or an \textit{attack} presentation. \textbf{Presentation attack instruments} (PAI) are those biometric characteristics or artificial objects used in presentation attacks. 

\paragraph{Impostor Attack Presentation}
Impostor attacks are typically generated from bona fide images of an iris. For example, attackers may have acquired the iris image of an individual with access to a system and wish to be granted access. One common impostor attack instrument is paper printouts of iris images, shown in Figure \ref{subfig:print}. Another impostor attack method is a replay attack, where bona fide iris images displayed on a screen are presented to the sensor. In general, formulating a successful Impostor Attack is more difficult than a concealer attack because you need the recognition software to determine you are a known individual, rather than the latter which requires the system to determine you are not a known individual.

\paragraph{Concealer Attack Presentation} 

 The most common Concealer Attack Presentation instrument is textured contact lenses. The texture on these lenses obscures substantial portions of the iris, preventing the iris recognition system from identifying the user. There are also contact lenses that are colored to alter the user's eye color appearance. Due to the wide range of manufacturers, all with unique designs, patterns and colors, these concealer attacks can be hard to distinguish from bona fide irises. In addition, the lenses may shift around on the eye such that different image captures of the same eye wearing the same textured contact may vary. Examples of different brands of textured contact lenses can be seen in Figure \ref{subfig:text}. The goal of this attack is simply to ensure anonymity of the user. It is also possible that attackers could use these textured contact lenses as an Impostor Attack, where a genuine iris texture is transcribed onto a lens. However, to our knowledge, this type of impostor attack has not yet been successfully demonstrated.

 Another possible concealer attack is 
 synthetic iris images that can imitate a bona fide iris pattern, Figure \ref{subfig:synth}.
 Synthetic samples pose a problem to PAD systems, 
 as even humans may have difficulty to distinguish between
 a (good) synthetic sample and a genuine iris..
 Synthetic iris images such as those found in CASIA-Iris-Synthetic (\cite{CASIA_IrisSynv4_URL}) and the work by \cite{Yadava,Yadav} can prove to be useful in training PAD systems to become more robust to unseen attacks. However, as these are generated images, the problem of how to present these samples to a sensor still exists. Thus, although synthetic irises can deceive software solutions, it is challenging to present this attack type to a sensor without having to use an aforementioned impostor attack such as printouts or a replay attack. 
 Especially as, or if, remote iris authentication becomes more widespread, synthetic presentation attacks will become viable and demand more attention.
 
 As shown in \cite{Trokielewicz2018}, the irises of deceased individuals can also be used as a presentation attack instrument. For some number of hours post-mortem, the texture of the iris remains intact enough to deceive an iris PAD system. Hypothetically, the post-mortem iris could be used as an impostor attack of the deceased individual. 
 However, more realistic is that someone may use an image of a post-mortem sample to hide their identity. Post-mortem iris samples closely resemble live irises in the early stages after death. Thus, detecting these samples in the wild may prove difficult.

\subsection{Error Rates}
Basic PAD-related error metrics include: \textbf{Attack Presentation Classification Error Rate (APCER)}, which refers to the proportion of \textit{attack presentations} incorrectly classified as \textit{bona fide presentations}; \textbf{Bona Fide Presentation Classification Error Rate (BPCER)}, which refers to the proportion of \textit{bona fide presentations} incorrectly classified as \textit{presentation attacks}; \textbf{Imposter Attack Presentation Match Rate (IAPMR)}, which refers to the proportion of impostor attack presentations that are successful, where the biometric reference for the targeted identity is matched (IAPMR is analogous to the false match rate (FMR) in identity verification); and \textbf{Concealer Attack Presentation Non-Match Rate (CAPNMR)}, which refers to the proportion of concealer attack presentations that are successful, where the biometric reference of the concealer is not matched (CAPNMR is analogous to the false non-match rate (FNMR) in identity verification).

\subsection{Acronyms}
Similar to~\cite{Czajka2018_ACM_Survey}, we summarize the meanings of several acronyms, used throughout the article:
BSIF: Binary Statistical Image Features, \cite{Kannala_2012_BSIF};
CNN: Convolutional Neural Network, \cite{Lecun_ProcIEEE_1998};
HoG: Histogram of oriented Gradients, \cite{Dalal_2005CVPR_HoG};
LBP: Local Binary Patterns, \cite{Ojala_ICPR_1994};
SID: Shift-Invariant Descriptor, \cite{Kokkinos_CVPR_2008}; and
SVM: Support Vector Machine, \cite{Boser_CLT_1992}.

\section{Databases To Support Iris PAD Research}

Since June 2018, seven new iris PAD datasets have been offered (excluding proprietary datasets). From the perspective of PAIs, three include textured contact lenses, two include post-mortem irises, and one includes prosthetic eyes. To provide a clearer and more direct comparison between datasets, we summarize the most important technical properties of the datasets in Tables~\ref{tab:datasets1} and \ref{tab:datasets2}. \par
There are several observations worth noting here. First, six out of seven newly collected datasets are static samples and only one database by \cite{Kinnison_ICB_2019} offers videos demonstrating iris/pupil dilation dynamics (but only live samples are included, without any spoof examples). This shows that static samples are still the most ubiquitous type of data used in iris PAD. Second, no images of irises printed out on paper and presented to the sensor are included in any of the new datasets, while in \cite{Czajka2018_ACM_Survey}, the most popular attack instrument in the datasets is iris printouts. Apparently, the current research focus has shifted from printouts to more challenging presentation attack instruments such as contact lens and postmortem irises. Third, \cite{Yadav2019} introduced images collected in both indoor (controlled) and outdoor (unconstrained) environment. The inclusion of images captured in unconstrained environments facilitates research for more robust algorithms that can be deployed on mobile devices.

\par
\cite{Czajka2018_ACM_Survey} also offered a review of the iris PAD competitions, which included the LivDet-Iris series in~\cite{LivDet2013, LivDet2015, LivDet2017}, as well as the MoblLive competition in~\cite{MoblLive2014}. No new competitions have been conducted since then.

\begin{table*}[htb!]
    \begin{center}
        \caption{Technical properties of datasets used in development of iris PAD methods.}
        \label{tab:datasets1}
        \footnotesize
        \begin{tabular}{lllll}
            {\bf Benchmark name}
            & {\bf Type}        
            & {\bf Wavelength}
            & {\bf Sensor(s)}
            & {\bf Spatial or temporal}\\
            {\bf [paper] } & {\bf of samples} & {\bf range} & {\bf used} & {\bf resolution} \\
            \hline
            \hline
            {ND WACV 2019},~\cite{Czajka2019} & CL & NIR & L4 & $640 \times 480$ px \\ 
            {ND Iris3D},~\cite{Fang2020a} & CL & NIR & A, L4 & $640 \times 480$ px  \\ 
            {Warsaw-BioBase-Postmortem-Iris-v2},~\cite{Trokielewicz2019_postv2} & PM & NIR & IS & $640 \times 480$ px \\ 
            & & VIS & TG3 & $640 \times 480$ px\\ 
            {Warsaw-BioBase-Postmortem-Iris-v3},~\cite{Trokielewicz2020_postv3} & PM & NIR & IS & $640 \times 480$ px \\ 
            & & VIS & TG3 & $640 \times 480$ px\\ 
            {Warsaw-BioBase-Pupil-Dynamics v3.0}, \cite{Kinnison_ICB_2019} & PD & NIR & SD & $768 \times 576$ px / 25 Hz \\
            {WVU Un-MIPA},~\cite{Yadav2019} & CL & NIR & BK,E,IS & $640 \times 480$ px \\
            \cite{Venkatesh2019} & PE & NIR & SD & $2448 \times 2048$ px  \\ 
            \hline
            \multicolumn{5}{l}{{\bf Type of samples:} CL - live + textured contact lenses; PE - live + prosthetic eyes; PM - post-mortem (cadaver) iris; PD - iris videos }\\
            \multicolumn{5}{l}{with pupil reaction to light stimuli. {\bf Wavelength:} NIR - near-infrared; VIS - visible light. {\bf Sensors:} A - IrisGuard AD 100; }
            \\
            \multicolumn{5}{l}{BK - IrisShield BK 2121U; E – CMITECH EMX-30; IS – IriShield MK2120U; L4 – LG4000; TG3 — Olympus TG-3; SD – Self-designed.}\\
        \end{tabular}
    \end{center}
\end{table*}

\begin{table*}
\centering
\caption{Subject breakdown information of datasets used in development of iris PAD methods.}
\label{tab:datasets2}
\footnotesize
\begin{tabular}{lrrrrrc}
\textbf{ Benchmark name }                           & \multicolumn{2}{c}{\textbf{ \# Distinct irises }} & \multicolumn{3}{c}{\textbf{ \# Samples }} & \multicolumn{1}{r}{\textbf{ Train/test }}  \\
\textbf{ [paper] }                                  & BF  & PA                                          & BF    & PA    & Total                     & \textbf{ split}                            \\ 
\hline\hline
ND WACV 2019,~\cite{Czajka2019}                     & 238 & 74                                          & 1,404 & 2,664 & 4,068                     & yes                                        \\
ND Iris3D,~\cite{Fang2020a}                         & 176 & 176                                         & 3,458 & 3,392 & 6,850                     & yes                                        \\
Warsaw-BioBase-Postmortem-Iris-v2 (NIR),~\cite{Trokielewicz2019_postv2} & 0   & 73                                          & 0     & 1,200 & 1,200                     & no                                         \\
Warsaw-BioBase-Postmortem-Iris-v2 (VIS),~\cite{Trokielewicz2019_postv2}   & 0   & 73                                          & 0     & 1,787 & 1,787                     & no                                         \\
Warsaw-BioBase-Postmortem-Iris-v3 (NIR),~\cite{Trokielewicz2020_postv3} & 0   & 42                                          & 0     & 1,094 & 1,094                     & no                                         \\
Warsaw-BioBase-Postmortem-Iris-v3 (VIS),~\cite{Trokielewicz2020_postv3}                                                   & 0   & 42                                          & 0     & 785   & 785                       & no                                         \\
 {Warsaw-BioBase-Pupil-Dynamics v3.0}, \cite{Kinnison_ICB_2019}                     & 84 & 0                                         & 117,117 & 0 & 117,117                    & no\\
WVU Un-MIPA,~\cite{Yadav2019}                     & 162 & 162                                         & 9,319 & 9,387 & 18,706                    & no                                         \\
\cite{Venkatesh2019}                        & 24  & 2                                           & 1,200 & 2,400 & 3,600                     & yes                                        \\ 
\hline
\multicolumn{7}{l}{BF = Bona Fide Samples, PA = Presentation Attack Samples}                                                                                                                    
\end{tabular}
\end{table*}

\section{Latest Proposed PAD Methodologies}
\subsection{Traditional Computer Vision-Based Methods}
Since 2018, most iris PAD research has shifted toward deep learning methods, but a few traditional computer vision-based methods have also been proposed. \cite{McGrath_2018_OSPAD} developed an open source PAD method based on 2D iris texture features for detecting textured contact lenses, available in both C++ and Python.
The method is an open source extension of the approach proposed by \cite{Doyle2015_NDCLD15}. Multi-scale BSIF are used as features and an ensemble of classifiers, including SVM, Multilayer Perceptron (MLP) and Random Forests (RF), is trained to make the prediction. 

This method obtains an accuracy on LivDet-Iris 2017 on par with that of the competition winner. Furthermore, the method uses a best guess about an iris location leveraging the fact that 
commercial iris sensors have the iris located near the center of the image. If an open-source segmentation software were included, the overall method should achieve better performance while remaining open-source. \par 

\cite{Venkatesh2019} designed a multi-spectral iris sensor with five frequency bands (800nm, 830nm, 850nm, 870nm, 980nm) to perform iris recognition and PAD. Several classes of feature extractors are used: texture-based (LBP and GLCM), image quality-based (BRISQUE), and spectral variation-based (spectral signature). Features across all descriptors and all wavelengths are fused using a weighted sum rule to perform the final classification. Since the main contribution of this paper is to propose a new sensor, data was collected specifically using the new sensor. The LBP-SVM achieves the best performance with 0\% BPCER and 5\% APCER. \par 
\cite{Czajka2019} proposed a photometric stereo-based 3D PAD method (OSPAD-3D). The method builds on the fact that when a bona fide iris is illuminated from opposite directions, the shadows observed in two images are minimal. However, for an iris wearing textured contact lens, significant differences in the shadows are observed in the images. Therefore, the reconstructed surface is relatively flat for bona fide irises but more irregular for irises with textured contact lens. Given a pair of masked iris images, OSPAD-3D estimates the surface normal vectors of the iris surface from photometric stereo, and the variance of the vectors' distances to the mean normal vector is computed as the PAD score. \par 
\cite{Wang2019} detect contact lenses by observing the change in curvature of the outer cornea surface caused by wearing contact lenses. For a bona fide iris, the curvature of each point on the cornea is basically unchanged, as it is a stable and detectable intrinsic property. After contact lenses are put on, however, the curvature of the outer cornea surface changes from a sphere to an ellipsoid, with the curvature large at the center and small at the margins. This method, unfortunately, is tested on a self-collected dataset. Although the authors report a 0\% error rate, no comparisons with other methods can be made. \par
Based on methods in \cite{McGrath_2018_OSPAD} and \cite{Czajka2019}, \cite{Fang2020a} proposed an OSPAD-fusion algorithm that fuses the 2D textural features (OSPAD-2D) and 3D photometric stereo features (OSPAD-3D). The authors identified that OSPAD-3D often fails to detect attack presentations of highly opaque contact lens, as they produce very little shadow, and OSPAD-2D often achieves a high APCER and low BPCER on unknown samples, so the samples marked as ``attack" by OSPAD-2D are usually correctly classified. Therefore, OSPAD-fusion employs a cascaded fusion algorithm to combine the strengths of both algorithms. The performance, as evaluated on \textit{NDCLD'15} and \textit{NDIris3D}, surpasses all other available open source iris PAD methods.

\subsection{Deep Learning-Based Methods}

With the rise in popularity of deep learning, it may come as no surprise that the field of iris PAD has followed that trend. 
There are multiple forms this application of deep learning may take. Proposed methodologies range from full end-to-end deep learning-based classification where the input is a raw or pre-processed image and the output is a PAD score or decision. Researchers have also shown that deep learning-based identity recognition models can be employed as feature extractors for iris PAD images. 
Finally, researchers have shown the power of adversarial networks in PAD. By training GANs to generate near-perfect synthetic iris images, the discriminator can be used to distinguish between bona fide samples and presentation attacks. 

The challenges that arise when using deep learning surround generalizability. Can we train models on one domain and expect it to perform reliably on another unseen domain? Deep learning has been shown to perform well when both training and testing data are from the same source(s). However, PAD has the property that we cannot predict what future attacks will look like, hence, methods need to be robust across domains.

\subsubsection{End-To-End CNNs}
\cite{Kuehlkamp2019} show how an ensemble of neural networks can be employed to transform BSIF representations of images into more discriminative features which enable the network to make stronger inferences. Predictions from the individual networks in the ensemble are then aggregated to output a decision. The cross-domain ability of this approach is shown and results that outperform the state-of-the-art are reported.
\cite{Yadav2019} propose a new PAD architecture DensePAD which utilizes the popular CNN architecture DenseNet. This proposed architecture takes normalized iris images of size $120 \times 160$ as input and outputs a decision as to whether the sample is bona fide or attack. Their paper addresses textured contact lenses in an uncontrolled and cross-sensor scenario, and presents good results on unseen types of textured contacts.
Good cross-dataset and cross-attack performance can also be seen in \cite{Hoffman2019,Hoffman2018}. In \cite{Hoffman2018} a CNN is employed to perform classification on patches of an iris region. The results suggest that textured contact lenses are the most difficult presentation attack to classify. This is later extended to \cite{Hoffman2019} which includes the ocular region. In that work, three CNNs are fused to generate decisions. Through analyzing the ocular region in conjunction with the iris, additional information can be attained that aids classification and strong cross-dataset performance is detailed. 

\cite{Chen2018} investigated whether information in the IrisCode (\cite{daugman2009iris}) can be useful for PAD. 
Three inputs are considered in this work as input to three CNNs. Un-normalized irises are found to allow more accurate detection, suggesting that liveness information may be lost during normalization. 
Textured contact lenses are again found to be more difficult to detect in comparison to paper printouts. The reason for this may be that the printed pattern is visible on the entire sample whereas the textured contact is only visible on the iris. \cite{Trokielewicz2018} employ a fine-tuned VGG-16 architecture to propose a method of iris PAD to detect post-mortem samples. This approach also provides analysis as to what features and regions the network deems most relevant to PAD classification by presenting the class activation maps. Results show a strong ability to detect post-mortem iris samples, but no cross-attack analysis is reported.

\subsubsection{Employing CNNs As Feature Extractors}

\cite{Nguyen2018} show how the combination of CNN based features for both global and local iris regions can result in more discriminative feature representations. To generate scores, SVMs are employed. This work explores feature-level fusion where the features are concatenated and passed to the SVM, as well as score-level fusion, where individual regions are passed to an SVM and then based on these scores another SVM is used to make the final decision. Various input types are also examined: three-channel gray images, three-channel Retinex images, and the fusion of both previous types into a third three-channel combination. The results show that this approach of feature extraction produced better results than using an end-to-end CNN and better results than all compared previous works. This method also shows resilience against unseen attack samples by presenting results on databases from the LivDet-Iris-2017 competition.

\subsubsection{Adversarial Learning}

Multiple modern approaches employed GANs for iris PAD. The logic for this is that if a discriminator network can be trained to accurately decide whether a synthetic sample is bona fide or not, then the same discriminator may be able to detect presentation attack samples that may exhibit non-natural artifacts such as a patterned iris or paper texture.

\cite{Yadava} hypothesized that these discriminator networks will generate a tight boundary around bona fide iris samples, such that any attack samples will fall outside this boundary. RaSGAN (\cite{jolicoeur2018relativistic}) is employed as the synthetic iris generator. The results show that the generated synthetic iris images are very similar to bona fide irises.
This work is extended in \cite{Yadav}. The relativistic discriminator is re-purposed for iris PAD. This one-class approach is outperformed by the compared approaches on previously seen attacks. However, results show that fine-tuning this discriminator network with a small number of presentation attack samples outperforms other methods on unseen attack types and hence has high generalization capabilities. 

\cite{Ferreira2019} also proposed the use of GANs to attain better generalization in iris PAD. The proposed methodology outlined that learning latent representations of images that are invariant to the presentation attack type yet still preserve information necessary to make the classification results in robust generalization against different attack types. However, the dataset used in this work is small and may not be representative of the individual domains. Their work concludes that the presented results outline that deep learning approaches with additional strategies will provide great development in iris PAD. 

\subsection{Hybrid Methods}

\cite{Yadav2018} combine the Haralick texture features in the multi-level Redundant Discrete Wavelet Transform (RDWT) domain with VGG features reduced by principal component anaysis. The two types of features are concatenated together as the input to a 3-layer MLP for binary classification as bona fide or attack. Experiments on the combined iris dataset proposed in~\cite{Kohli2016DESIST} show that the proposed fusion method outperforms Haralick features or VGG features alone. The method also achieves better results than several baselines including LBP, WLBP, and DESIST. \par
Building upon six traditional features (BSIF, LBP, CoA-LBP~\cite{Nosaka_2012_CoALBP}, HoG, DAISY~\cite{Tola_2010_Daisy}, and SID) and one deep feature extracted by the first seven layers of VGG, \cite{Poster2018} propose to learn the best subset of features through group sparsity. 
Group dropout operation is used to avoid excessive reliance on certain features 
and a novel group sparsity-based regularization strategy is adopted to mitigate overfitting. The authors evaluate the proposed method on NDCLD'13, IIITD (Cogent and Vista), and Clarkson LivDet-Iris 2013 datasets. On NDCLD'13 and IIITD, the method outperforms the state-of-the-art method. On Clarkson LivDet-Iris 2013, the method outperforms the winner of the competition. \par
\cite{Choudhary2020} performs a score-level fusion of data-driven features learned from a customized Densenet121 architecture and the same set of handcrafted features as in~\cite{Poster2018}. The score-level fusion is guided by a Friedman test which identifies the top $k$ features to include in the fusion. The authors accommodated a wide range of experimental setups including intra-sensor, inter-sensor, and combined-sensor tests and with both textured and soft contact lens, on several benchmark datasets: NDCLD'13, IIITD (Cogent and Vista), and Clarkson LivDet-Iris 2017. The method further outperforms~\cite{Poster2018} and all previous state-of-the-art methods in almost all experiments. \par

\section{Performance of Methods}
\label{sec:performance}

This section summarizes the performance comparison of the PAD methods covered in this paper. We observe that most methods differ in datasets, train/test splits, and evaluation metrics. Therefore, we adopt the following strategy when comparing their performance: for methods that do not have source codes available, we group them by the datasets and train/test split used and report the results as in the original papers.
For open source methods, we attempt to compare all methods whose source codes can be obtained from the internet or through contacting the authors.

\subsection{Comparison of Methods Grouped by Datasets}
In~\cite{Mandalapu2019}, the authors compared five different PAD methods on four different datasets whose PAIs include printouts and patterned contact lenses. All five methods are traditional vision-based methods, where the feature extractors are adopted from previous PAD papers. Through extensive experiments, the authors discovered that the fusion of texture (BSIF) and image quality (BRISQUE) leads to the best performance for unknown attacks. In contrast, when all attacks and sensors are included in the development of the PAD algorithm, color adaptive quantized patterns (CAQP) achieves the best performance. Furthermore, the experiments in the paper demonstrate that the fusion of multiple high performing features generally leads to higher accuracy. \par 
In~\cite{Choudhary2020}, the authors compared their method against~\cite{Poster2018} and~\cite{Yadav2018} on a wide range of datasets: NDCLD'13, IIITD (Cogent and Vista), Clarkson Livdet 2017. Both intra-dataset and cross-dataset experiments are performed. \cite{Choudhary2020} ranked first in most cross-dataset scenarios and achieved low error rates in intra-dataset settings as well. The method proposed by \cite{Poster2018} is the next best approach with consistent performance across settings. \cite{Yadav2018} achieve near-perfect performance on NDCLD'13 intra-dataset test, but perform less well on other datasets and always ranked last in cross-dataset settings, indicating its inability to generalize well to unknown types of attacks. \par 
Other methods available for comparison are those from the same family of work. \cite{Yadav2019} compared against~\cite{Yadav2018} and achieved better results. 
\cite{Hoffman2019} showed improved performance over \cite{Hoffman2018} on the same datasets. In those cases, however, no comparisons with other methods are offered. For other papers, either no performance comparisons are provided or different train/test split are used for reporting the performance. This makes it challenging for the community to compare methods when multiple papers claim to achieve state-of-the-art results.

\subsection{Comparison of Open Source Methods}
The only comprehensive comparison of open source methods known to us is~\cite{Fang2020a}.
To the best of our knowledge, no new open source methods have been released since that paper. Three modern publications in this paper,~\cite{McGrath_2018_OSPAD, Czajka2019, Fang2020a}, along with three older ones are included in the comparison using the same protocol. The authors found that the PAD method based on photometric stereo features~\cite{Czajka2019} generalize better to attacks of contact lens of unknown textures, while the BSIF texture-based PAD method~\cite{McGrath_2018_OSPAD} performs better in closed-set scenarios. Experiments also show that the fusion method~\cite{Fang2020a} outperforms the other two methods in both known and unknown settings. This finding agrees with~\cite{Mandalapu2019}. 

\section{Future Research Directions}

\paragraph{Standardized evaluation platform} \cite{Czajka2018_ACM_Survey} reported in 2018 that the only available iris PAD evaluation platform is the LivDet-Iris series, and there were no platforms for asynchronous evaluation of iris PAD algorithms. To the best of our knowledge, we still do not have such a platform. As observed in Section~\ref{sec:performance}, fair comparison between methods, especially those without source codes, is still very challenging. A standardized, accessible, and fair platform for PAD evaluation will facilitate the comparison between PAD methods. 

\paragraph{Fairness in iris PAD}
\cite{Fang2020} study gender bias in iris PAD. Three different experimental classifiers are examined and for all three it shows that the error rates for males are lower than for females. To our knowledge, this is the first work examining demographic bias in PAD, for any modality. The authors note possible future extensions to examining bias in eye color. There could also be room to investigate the accuracy of iris PAD across race. There has been much work on bias in facial recognition systems such as in \cite{Albiero_2020_WACV}. 
Although considerably less demographic information is available in an iris sample, it would still be a worthwhile endeavor to investigate biases, seeing as \cite{Fang2020} concluded that females seem to be significantly less protected by iris PAD systems.

\paragraph{Making methods open source} 
\cite{Fang2020a} lists six iris PAD methods that were either publicly available or available by contacting the authors. 
The field would benefit from more methods becoming open source, so that proposed methodologies can be easily benchmarked against the current state-of-the-art. Open-sourcing your code also furthers reproducibility, enabling researchers to make modifications and improvements directly rather than having to re-implement methods based on published descriptions, thus decreasing the time required to run experiments. 

\paragraph{Generalization to unknown attack types} 
The ability to be robust to unseen attack types is crucially important. Attackers are continually developing new attack methodologies to circumvent iris PAD systems. In the future, the main goal of iris PAD should be the ability to detect unseen attack types while maintaining high accuracy on known attacks. In the work by \cite{Yadava}, a tight boundary around bona fide samples is generated using a GAN. It showed increased accuracy against unseen attacks; however, the compared work outperformed the GAN approach on known attacks. It seems from previous works that it is a trade-off between exceptional performance on known attacks but poorer performance on unseen attacks, or good performance on unseen attacks but worse performance on known attacks. One possible future direction could be trying to bridge the gap between known attacks and unseen attacks. Is there a way to more precisely model bona fide irises such that attack samples can be easily distinguishable?

\section{Summary}

This paper summarizes recent advancements in iris PAD since the release of the survey by \cite{Czajka2018_ACM_Survey}. New publicly available datasets are outlined and described. We show that modern methodologies can be grouped into one of three sets: traditional hand-crafted feature extraction and classification, deep learning-based approaches, and hybrid approaches that use both traditional and deep-learning in conjunction.
Commentary is provided on the performance of the studied methods in comparison to one another. Finally, possible future research directions are given to help inspire new works.


{\small
\bibliographystyle{IEEE}
\bibliography{refs}
}
\end{document}